\documentclass[11pt]{article}

\usepackage{acl}

\usepackage[utf8]{inputenc}
\usepackage[T1]{fontenc}
\usepackage{times}

\usepackage{latexsym}
\usepackage{microtype}
\usepackage{inconsolata}

\usepackage{amsmath}
\usepackage{amsfonts}

\usepackage{graphicx}
\usepackage{booktabs}
\usepackage{tabularx}
\usepackage{longtable}
\usepackage{multirow}
\usepackage{array}
\usepackage{makecell}

\usepackage{xcolor}
\usepackage{url}
\usepackage{seqsplit}
\usepackage{comment}
\usepackage{enumitem}
\usepackage{placeins}
\usepackage{float}

\usepackage{subcaption}

\usepackage{tikz}
\usetikzlibrary{arrows.meta,positioning,fit,calc,backgrounds}

\usepackage{todonotes}

\usepackage{CJKutf8}
\newcommand{\zh}[1]{\begin{CJK*}{UTF8}{gbsn}#1\end{CJK*}}

\definecolor{freqcolor}{HTML}{EF553B}
\definecolor{lexcolor}{HTML}{00CC96}
\definecolor{mlmcolor}{HTML}{AB63FA}
\definecolor{surpcolor}{HTML}{FFA15A}
\definecolor{semcolor}{HTML}{B6E880}
\definecolor{cogcolor}{HTML}{636EFA}

\title{UOL@IDEM at BEA 2026 Shared Task 1:\\
Neural Fusion and Feature-Rich Modeling for L1-Aware Vocabulary Difficulty Prediction}

\author{
Nouran Khallaf$^{1,2}$ \quad Serge Sharoff$^{1}$\\
$^{1}$Centre for Translation, Localisation and Interpreting Studies,\\
School of Languages, Cultures and Societies, University of Leeds, UK\\
$^{2}$Alexandria University, Egypt\\
\texttt{\{n.khallaf,s.sharoff\}@leeds.ac.uk}
}

\begin{document}
\maketitle

\begin{abstract}
This paper describes UOL@IDEM's closed-track submission to the BEA 2026 shared task on L1-aware vocabulary difficulty prediction\footnote{\url{https://github.com/Nouran-Khallaf/UoL-IDEM-BEA2026-Vocabulary-Difficulty-Prediction}}. We model the task as regression and train separate systems for Spanish, German, and Mandarin Chinese\footnote{Below we use \emph{Chinese} for brevity.}. Our system combines multilingual contextual representations with engineered features capturing frequency, surface form, retrieval evidence, semantic alignment, cognate similarity, and masked-language-model predictability. Development results show consistent gains over the official closed-track baselines, with sentence-embedding encoders such as BGE-M3, multilingual E5, and LaBSE performing best. Official submissions achieve RMSE scores of 1.132, 1.037, and 0.891 for Spanish, German, and Chinese, respectively. Feature analysis identifies frequency as the most stable predictor, while contextual predictability, form similarity, retrieval, and semantic features provide complementary L1-sensitive signals. Error analysis shows strong ranking performance but weaker calibration for the easiest items, which are often overpredicted.
\end{abstract}

\section{Introduction}

Text complexity and readability assessment are central to educational NLP, language learning, and text adaptation. A key component is lexical complexity prediction~(LCP), which estimates how difficult a word or multi-word expression is likely to be for a target reader \citep{shardlow-etal-2021-semeval,shardlow-etal-2022-complex,north-etal-2023-overview}. Lexical difficulty is shaped by frequency, word length, morphology, polysemy, context, and reader background, and supports applications such as readability assessment, simplification, tutoring, and machine translation \citep{shardlow-etal-2022-complex,north-etal-2023-overview,ohuoba-etal-2024-quantifying}.

The BEA 2026 Shared Task extends LCP by making vocabulary difficulty explicitly L1-aware. In this setting, the difficulty of an English target word depends not only on the word itself, but also on the context of its use as well as on its relationship to the learner’s first language through translation and transfer cues. As a result, the same English word may have different difficulty profiles for Spanish, German, and Chinese speakers.  More specifically, the task uses an extended version of the Knowledge-based Vocabulary Lists (KVL) resource, pairing English target items with multilingual prompts and psychometrically calibrated GLMM-based difficulty scores \citep{skidmore-etal-2025-transformer}.

In the closed track, systems train separate models for each L1 using only the shared-task resources \citep{felice-skidmore-2026-bea-shared-task, bea2026task}. We therefore model the task as three parallel prediction settings: Spanish$\rightarrow$English, German$\rightarrow$English, and Chinese$\rightarrow$English.

Our approach experiments with mid-sized text encoders (of up to 560M parameters) enriched with engineered features, such as lexical frequency, orthographic clues, contextual retrieval, cognate-like similarity, semantic-domain alignment, and masked-language-model predictability. We evaluate several encoders within a neural fusion framework and compare full-feature and reduced-feature variants.

Our experiments show three main findings. First, neural fusion improves over the official closed-track baselines across all three languages, improving RMSE by up to 0.21--0.26. Second, sentence-embedding-oriented multilingual models, especially BGE-M3 \citep{chen-etal-2024-m3}, multilingual E5 \citep{wang2024multilinguale5}, and LaBSE \citep{feng-etal-2022-language}, are more effective than plain transformer baselines such as mBERT and XLM-R. Third, feature and error analyses show that frequency is the most stable predictor contributing above the encoder baseline, while semantic, cognate, retrieval, and MLM-based cues can provide complementary L1-sensitive information.

\section{Data and Preprocessing}
\label{sec:data_preprocessing}

Data exploration showed that \texttt{L1\_source\_word} often contains noisy forms that require language-aware normalisation before feature extraction. We address lexical variants, negative constraints, morphological fragments, negated instructional patterns, explanatory glosses, and punctuation. For example, multiple alternatives are split and tracked with a variant-count feature (\textit{Korridor, Flur} $\rightarrow$ \textit{Korridor}); exclusion notes are removed and encoded as binary features (\textit{Flugzeug (nicht: ...)} $\rightarrow$ \textit{Flugzeug}); truncated German compounds are normalised to stems (\textit{Regierungs-} $\rightarrow$ \textit{Regierungs}); Spanish instruction-like phrases such as \textit{no es} and \textit{la respuesta no termina en} are stripped to isolate the lexical item; and Chinese punctuation, leading glosses, and parenthetical notes are pruned where appropriate (\zh{（美）走廊，过道} $\rightarrow$ \zh{走廊}). To make preprocessing traceable, we keep the original source-word string and use a cleaned lexical form for feature extraction. When an English target contains multiple candidates, we select the candidate that appears in the L1 context.
\section{Methodology}
We formulated the task as supervised regression. Each instance contained an L1-specific context, an L1 source word, an English target word, and a continuous difficulty score. Models used textual input and engineered features capturing complementary cues: frequency, lexical form, retrieval evidence, MLM and surprisal estimates, semantic-domain information, and cognate-based similarity.

We compared three integration strategies: (1) textualised feature regression, which renders selected features as part of the input prompt; (2) multi-stage late fusion, which combines frozen encoder embeddings with downstream regressors; and (3) neural fusion, which jointly optimises contextual representations and structured tabular features. Because neural fusion performed best in the initial comparison, we used it as the main setting.

\subsection{Engineered features}
\label{secEngineered}
In addition to BERT-derived text representations, we designed a set of engineered features to capture lexical difficulty, cross-lingual retrievability, contextual predictability, semantic compatibility, and source-word ambiguity. We group these features into frequency-based, lexical, retrieval-based, masked-language-model and surprisal, semantic, and cognate-based features. Full feature descriptions are provided in Appendix~\ref{app:features}, Table ~\ref{tab:appendix_engineered_features}.
\paragraph{Frequency-based features.}

We combine three complementary frequency resources. KELLY \citep{kilgarriff-etal-2014-kelly} is derived from curated Web-corpora and provides pedagogically oriented indicators, including frequency rank, percentile, and CEFR-aligned labels, which approximate learner vocabulary progression\citep{council-europe-2001-cefr}.
The \texttt{wordfreq} Python package provides general corpus-based estimates, including raw frequency, Zipf frequency, frequency percentile, and inverse-frequency cost \citep{speer-2022-wordfreq,sharoff-2017-frequen_ruwac}. The SUBTLEX list is derived from subtitles and provides frequency and contextual-diversity measures, reflecting both occurrence frequency and exposure across everyday contexts \citep{brysbaert-new-2009-subtlex,vanheuven-etal-2014-subtlexuk}.  Combined, these features model familiarity from pedagogical, general-corpus, and everyday-exposure perspectives, providing a strong baseline signal for vocabulary difficulty prediction \citep{shardlow-etal-2021-semeval,shardlow-etal-2022-complex,north-etal-2023-overview}.

\paragraph{Surface-form and morphosyntactic features.}
We extract surface-form and morphosyntactic features from the English target and cleaned L1 source word. These include target/source length, target syllable count, clue informativeness, and  hidden clue characters. We also encode POS information, using the dataset-provided target POS and spaCy-derived source POS \citep{honnibal_spacy_2020}. Alongside these features, we also retain a set of preprocessing-derived indicators introduced in Section~\ref{sec:data_preprocessing}, including signals for exclusion notes, alternative candidate forms, and related counts.

\paragraph{Retrieval-based features.} We derive retrieval-based features by treating the task as context-guided cross-lingual retrieval of the English target. We build a fixed English candidate bank from possible target words, encode candidates with mBERT \citep{devlin-etal-2019-bert}, and map them into a shared multilingual embedding space. For each instance, the L1 context and source word are encoded as a source-side representation and matched against this bank.
Candidates are reranked using the available constraints: initial-letter clue, target length, and part of speech. Matching candidates are promoted, while mismatches are downweighted or excluded. We then extract features describing gold-target support, including rank before and after reranking, cosine and reranked scores, top-candidate scores, margins, nearest-neighbour confidence and entropy, gold-target probability measures, and the number of valid candidates. The gold target's retrieval position is encoded as its rank when retrieved and as 0 otherwise. These features capture how recoverable the target is from the L1 context and source word.

\paragraph{Masked language model and surprisal features.}
We use the MLM head of mBERT \citep{devlin-etal-2019-bert} to estimate how predictable the English target is in context. Unlike retrieval features, which compare the target with an external candidate bank, MLM features come from the output distribution at the masked target position. We apply target constraints, especially initial letter and length, so the scores reflect both contextual fit and target-form compatibility. Features include gold-target log probability and rank, masked-token entropy, top-1 indicators, prediction margins, target--prediction embedding similarity, and related confidence measures.

Because targets may contain multiple wordpieces, probability-based quantities are computed at the subword level and averaged across the target span. For a target sequence $t_{1:m}$, we use three surprisal variants. Simultaneous masking replaces the full target span with \texttt{[MASK]}:

\begin{equation}
s_{\mathrm{masked}}(t_{1:m}) =
- \frac{1}{m} \sum_{j=1}^{m}
\log P\!\left(t_j \mid \mathbf{c}_{\mathrm{mask}}\right).
\end{equation}

Pseudo-log-likelihood (PLL) masks one target wordpiece at a time while keeping the remaining pieces visible \citep{salazar-etal-2020-masked}:
\begin{equation}
s_{\mathrm{PLL}}(t_{1:m}) =
- \frac{1}{m} \sum_{j=1}^{m}
\log P\!\left(t_j \mid \mathbf{c}^{(j)}\right),
\end{equation}
where $\mathbf{c}^{(j)}$ masks only token $t_j$, allowing each token to be predicted with access to the other target subwords.

Finally, chain-rule surprisal scores the target incrementally from left to right:
\begin{equation}
s_{\mathrm{chain}}(t_{1:m}) =
- \frac{1}{m} \sum_{j=1}^{m}
\log P\!\left(t_j \mid \mathbf{c}, t_{1:j-1}\right).
\end{equation}
These measures capture contextual predictability under full-span masking, single-token masking, and sequential prediction.
\paragraph{Semantic features.}
We derive semantic-domain features from the UCREL Semantic Analysis System (USAS), which assigns lexical items to 21 major discourse fields and 232 finer-grained categories \citep{rayson-etal-2004-usas}. For each item, we annotate the L1 source word and English target with all available USAS tags. We include overlap features for identical tags, shared fine-grained prefixes, and shared major fields. We also compute weighted and unweighted tag entropy to capture semantic ambiguity, with higher entropy indicating less stable interpretation.

To capture relatedness beyond exact overlap, we compute a semantic-shift score. Let $S$ and $T$ be the USAS tag sets for the L1 source and English target. For each source--target tag pair, we assign a graded relation score based on the strongest relation: exact tag overlap, shared fine-grained prefix, shared major field, related major fields, or unrelated fields. Item-level soft similarity is computed over the full tag sets, and semantic shift is defined as its complement:

\begin{equation}
\mathrm{shift}_{\mathrm{sem}}(S,T)
= 1 - \mathrm{sim}_{\mathrm{soft}}(S,T).
\end{equation}
Lower values indicate stronger semantic alignment; higher values indicate larger semantic-domain shift. For example, \textit{obra} (`literary work' in Spanish) has the semantic tags of \texttt{A1.1.1|I3.1}, which do not overlap with \textit{comedy} \texttt{E4.1+|K4|Q4.3}. Still, both refer to related fields, giving soft similarity 0.20 and semantic shift 0.80. Coverage gaps are handled using fixed penalties for unmatched \texttt{Z99} cases. Appendix~\ref{app:usas-shift} ,Table~\ref{tab:soft-usas-shift-examples} gives further examples with full USAS category names.

\paragraph{Cognate features.}
We derive cognate-oriented features to capture lexical-transfer signals between the L1 source word and the English target. These include orthographic similarity, multilingual embedding cosine similarity, Levenshtein edit-distance similarity, character $n$-gram overlap, and prefix/suffix overlap. We also add lexical-relation features from CogNet, a multilingual cognate database \citep{batsuren-etal-2019-cognet,batsuren-etal-2021-cognet}. Because the task covers English, Spanish, German, and Chinese, we restrict CogNet to entries involving these languages. These features are expected to be most informative for Spanish--English and German--English, while also testing whether cognate relations provide useful signals for Chinese--English.

\subsection{Model Architectures}

\paragraph{Textualised Feature Regression.}
The main text fields and selected engineered features are rendered as text, encoded with a multilingual transformer, and passed to a regression head. This allows feature--context interactions to be modelled through self-attention.

\paragraph{Multi-stage Late Fusion.}
Here the available text fields are concatenated and encoded with a frozen multilingual sentence encoder to obtain dense representations $\mathbf{h} \in \mathbb{R}^d$. These are used alone or concatenated with engineered features $\mathbf{x}$ as input to Ridge, GBDT, XGBoost, and SVR regressors. Their predictions are combined through ensemble-style aggregation \citep{van-der-laan-etal-2007-super}, following feature-based regression approaches in quality estimation and related prediction tasks \citep{rios-sharoff-2016-language}.

\paragraph{Neural Fusion.}
This architecture jointly learns from contextual text representations and structured tabular features. For each instance, the input sequence $S_i$ concatenates \texttt{L1\_context}, \texttt{L1\_source\_word}, \texttt{en\_target\_word}, and, in some configurations, an additional target clue using [SEP] tokens.

A multilingual transformer $f_\theta$ encodes $S_i$, and its token-level representations are pooled into a text vector:
\begin{equation}
    \mathbf{h}_i = \mathrm{Pool}(f_\theta(S_i)).
\end{equation}

In parallel, a feed-forward tabular branch maps the engineered feature vector $\mathbf{x}_i$ into a dense representation $\mathbf{x}'_i$. The two representations are concatenated:
\begin{equation}
    \mathbf{z}_i = [\mathbf{h}_i ; \mathbf{x}'_i],
\end{equation}
and passed to an MLP to predict the difficulty score:
\begin{equation}
    \hat{y}_i = \mathrm{MLP}(\mathbf{z}_i).
\end{equation}

Text representations vary by model family: neural fusion uses attention pooling over final hidden states, textualised regression uses the standard $[\mathrm{CLS}]$ representation, and late fusion uses dense sentence embeddings, optionally followed by PCA.
\begin{table*}[t]
\centering
\small
\begin{tabular}{llccccc}
\toprule
\textbf{Language} & \textbf{System} & \textbf{RMSE$\downarrow$} & \textbf{$\Delta$ RMSE vs Closed $\uparrow$ }& \textbf{Pearson $\uparrow$} & \textbf{Spearman $\uparrow$} & \textbf{Kendall $\uparrow$}\\
\midrule
es & Baseline (closed) & 1.3570 & -- & 0.7480 & -- & -- \\
es & LaBSE & 1.1732 & 0.1838 & 0.8116 & 0.8136 & 0.6189 \\
es & XLM-R-large & 1.1189 & 0.2381 & 0.8291 & 0.8340 & 0.6416 \\
es & mBERT & \colorbox{orange!20}{1.3269} & 0.0301 & 0.7415 & 0.7459 & 0.5475 \\
es & multilingual-E5 & 1.1134 & 0.2436 & 0.8301 & 0.8335 & 0.6386 \\
es & BGE-M3 & \colorbox{green!30}{1.0952} & \colorbox{green!30}{0.2618} & \colorbox{green!30}{0.8324} & \colorbox{green!30}{0.8373} & \colorbox{green!30}{0.6473} \\
es & Late fusion & 1.2667 & 0.0903 & 0.7513 & 0.7583 & 0.5620 \\
es & Textualised Feature & 1.1261 & 0.2309 & 0.8214 & 0.8239 & 0.6324 \\
\midrule
de & Baseline (closed) & 1.3280 & -- & 0.7530 & -- & -- \\
de & LaBSE & 1.1568 & 0.1712 & 0.8015 & 0.8139 & 0.6208 \\
de & XLM-R-base & 1.2211 & 0.1069 & 0.7885 & 0.8058 & 0.6093 \\
de & mBERT & \colorbox{orange!20}{1.2693} & 0.0587 & 0.7481 & 0.7640 & 0.5665 \\
de & multilingual-E5-large & \colorbox{green!30}{1.0873} & \colorbox{green!30}{0.2407} & \colorbox{green!30}{0.8234} & \colorbox{green!30}{0.8414} & \colorbox{green!30}{0.6446} \\
de & BGE-M3 & 1.1457 & 0.1823 & 0.8135 & 0.8260 & 0.6284 \\
de & Late fusion & \colorbox{orange!20}{1.2627} & 0.0653 & 0.7219 & 0.7365 & 0.5389 \\
de & Textualised Feature & 1.0961 & 0.2319 & 0.8252 & 0.8417 & 0.6466 \\
\midrule
cn & Baseline (closed) & 1.1750 & -- & 0.7360 & -- & -- \\
cn & LaBSE & 0.9937 & 0.1813 & 0.8179 & 0.8220 & 0.6343 \\
cn & XLM-R-large & \colorbox{orange!20}{1.2552} & \colorbox{orange!20}{-0.0802} & 0.6704 & 0.6699 & 0.4906 \\
cn & mBERT & 1.1227 & 0.0523 & 0.7673 & 0.7737 & 0.5868 \\
cn & multilingual-E5 & \colorbox{green!10}{0.9726} & \colorbox{green!10}{0.2024} & \colorbox{green!10}{0.8340} & \colorbox{green!10}{0.8387} & \colorbox{green!10}{0.6493} \\
cn & BGE-M3 & \colorbox{green!30}{0.9681} & \colorbox{green!30}{0.2069} & \colorbox{green!30}{0.8351} & \colorbox{green!30}{0.8428} & \colorbox{green!30}{0.6568} \\
cn & Late fusion & 1.0929 & 0.0821 & 0.7623 & 0.7722 & 0.5775 \\
cn & Textualised Feature & \colorbox{green!10}{0.9750} & \colorbox{green!10}{0.2000} & \colorbox{green!10}{0.8283} & \colorbox{green!10}{0.8362} & \colorbox{green!10}{0.6459} \\
\bottomrule
\end{tabular}

\caption{Main development-set results [dev] for the closed track. $\Delta$RMSE is computed as $\mathrm{RMSE}_{\text{baseline}} - \mathrm{RMSE}_{\text{system}}$, so higher values indicate larger improvements. \colorbox{green!30}{\strut best}, \colorbox{green!10}{\strut near-best}, \colorbox{orange!20}{\strut close to baseline}.}
 
\label{tab:dev_main}
\end{table*}
\subsection{Experimental Setup and Optimisation}

We train separate models for Spanish, German, and Chinese. For each language, we evaluate multilingual PLMs, including XLM-R and mBERT \citep{conneau-etal-2020-unsupervised,devlin-etal-2019-bert}, and sentence-oriented multilingual encoders, including LaBSE, Multilingual-E5, and BGE-M3 \citep{feng-etal-2022-language,wang2024multilinguale5,chen-etal-2024-m3}.

All experiments use five-fold cross-validation, with out-of-fold predictions used for training-set evaluation and held-out embeddings used for late fusion to avoid leakage. Models are optimised with AdamW and early stopping on validation RMSE, with differential learning rates, warmup, gradient clipping, and mixed precision. We test both mean squared error and Huber loss, the latter reducing the influence of large residuals; details are given in Appendix Eq.~\ref{eq:huber-loss} and Appendix~\ref{app:neural-fusion-settings_sec},Table~\ref{tab:neural-fusion-settings}. RMSE is the primary metric used by the shared task organisers, while Pearson, Spearman, and Kendall's $\tau$ used to assess fit and ranking quality of each method. 
\section{Results}

Table~\ref{tab:dev_main} reports the main development-set results for the closed track, comparing Textualised Feature Regression, Multi-stage Late Fusion, and Neural Fusion. Across all three languages, the best systems substantially outperform the official closed-track baselines, with RMSE gains of 0.2618 for Spanish using BGE-M3, 0.2407 for German using multilingual-E5-large, and 0.2069 for Chinese using BGE-M3.

The results show a clear advantage for sentence-embedding-oriented multilingual encoders. 
BGE-M3 and multilingual-E5-large, both based on XLM-RoBERTa-large, are consistently stronger, while LaBSE, a multilingual BERT-based sentence embedding model, remains competitive. 
This suggests that the main advantage comes not only from the underlying multilingual backbone, but also from training the model specifically for sentence-level cross-lingual representation. 
In contrast, the standard token encoders are weaker, mBERT is generally less competitive, and XLM-R is less stable, performing reasonably well for Spanish and German but falling below the baseline for Chinese.

For the Late Fusion and Textualised Feature Regression settings, we therefore used the two strongest sentence-embedding encoders, BGE-M3 and multilingual-E5-large, and report the best development-set result for each language. 
The modelling strategy also has a clear effect. Late Fusion performs worse than both Textualised Feature Regression and Neural Fusion, indicating that simply combining frozen sentence embeddings with classical regressors is not the most effective way to use these representations. In contrast, Neural Fusion benefits from combining encoder representations with engineered features in a more integrated way. Since the best development results are obtained with Neural Fusion using BGE-M3 or multilingual-E5-large, we use this setting for the subsequent feature-ablation and feature-importance analyses.

\subsection{Feature importance and ablations}

\begin{table*}[t]
\centering
\scriptsize
\setlength{\tabcolsep}{4pt}
\resizebox{\textwidth}{!}{%
\begin{tabular}{llcccccccc}
\toprule
\multirow{2}{*}{\textbf{Language}} & \multirow{2}{*}{\textbf{Feature setting}} & \multicolumn{4}{c}{\textbf{OOF}} & \multicolumn{4}{c}{\textbf{Dev}} \\
\cmidrule(lr){3-6} \cmidrule(lr){7-10}
 &  & \textbf{RMSE} & \textbf{Pearson} & \textbf{Spearman} & \textbf{Kendall} & \textbf{RMSE} & \textbf{Pearson} & \textbf{Spearman} & \textbf{Kendall} \\
\midrule
\multirow{10}{*}{es}
 & All features & \colorbox{green!10}{1.0988} & \colorbox{green!30}{0.8172} & \colorbox{green!30}{0.8149} & \colorbox{green!30}{0.6223} & \colorbox{green!10}{1.0952} & \colorbox{green!30}{0.8324} & \colorbox{green!30}{0.8373} & \colorbox{green!30}{0.6473} \\
 & Top-1/group (Kendall) & 1.1429 & 0.7931 & 0.7900 & 0.5954 & 1.1256 & 0.8244 & 0.8298 & 0.6378 \\
 & Top-3/group (Kendall) & 1.1221 & 0.8061 & 0.8036 & 0.6089 & 1.1143 & 0.8308 & 0.8347 & \colorbox{green!10}{0.6452} \\
 & Frequency & \colorbox{green!30}{1.0987} & \colorbox{green!10}{0.8122} & \colorbox{green!10}{0.8084} & \colorbox{green!10}{0.6157} & \colorbox{green!30}{1.0855} & \colorbox{green!10}{0.8320} & 0.8350 & 0.6450 \\
 & Surface-form & 1.1343 & 0.8036 & 0.8013 & 0.6076 & 1.1739 & 0.8216 & 0.8238 & 0.6315 \\
 & MLM & \colorbox{orange!20}{1.2928} & 0.7257 & 0.7174 & 0.5274 & 1.1099 & 0.8295 & 0.8297 & 0.6404 \\
 & Surprisal & 1.1268 & 0.7997 & 0.7966 & 0.6022 & \colorbox{green!10}{1.0973} & 0.8304 & \colorbox{green!10}{0.8354} & 0.6440 \\
 & Semantic & 1.1276 & 0.8043 & 0.8029 & 0.6099 & 1.1209 & 0.8269 & 0.8275 & 0.6381 \\
 & Cognate & 1.1281 & 0.8059 & 0.8036 & 0.6104 & 1.1235 & 0.8237 & 0.8289 & 0.6386 \\
 & Features only & \colorbox{orange!20}{1.3248} & \colorbox{orange!20}{0.7027} & \colorbox{orange!20}{0.6927} & \colorbox{orange!20}{0.5030} & \colorbox{orange!20}{1.3274} & \colorbox{orange!20}{0.7136} & \colorbox{orange!20}{0.7079} & \colorbox{orange!20}{0.5130} \\
\midrule
\multirow{10}{*}{de}
 & All features & 1.0761 & 0.7987 & 0.8036 & 0.6080 & \colorbox{green!10}{1.0873} & 0.8234 & \colorbox{green!30}{0.8414} & \colorbox{green!10}{0.6446} \\
 & Top-1/group (Kendall) & 1.0538 & \colorbox{green!30}{0.8210} & \colorbox{green!30}{0.8254} & \colorbox{green!30}{0.6313} & 1.1056 & \colorbox{green!10}{0.8250} & 0.8398 & 0.6428 \\
 & Top-3/group (Kendall) & \colorbox{green!10}{1.0476} & 0.8096 & 0.8135 & 0.6189 & 1.1306 & 0.8123 & 0.8277 & 0.6304 \\
 & Frequency & \colorbox{green!30}{1.0452} & \colorbox{green!10}{0.8147} & \colorbox{green!10}{0.8201} & \colorbox{green!10}{0.6256} & 1.0941 & 0.8235 & 0.8381 & 0.6413 \\
 & Surface-form & 1.0805 & 0.8060 & 0.8125 & 0.6168 & 1.1073 & 0.8173 & 0.8335 & 0.6361 \\
 & MLM & 1.0778 & 0.8028 & 0.8088 & 0.6129 & 1.1255 & 0.8132 & 0.8280 & 0.6285 \\
 & Surprisal & 1.0831 & 0.7963 & 0.8025 & 0.6060 & \colorbox{green!30}{1.0861} & \colorbox{green!30}{0.8261} & \colorbox{green!10}{0.8403} & \colorbox{green!30}{0.6454} \\
 & Semantic & 1.0744 & 0.7975 & 0.8036 & 0.6075 & 1.1343 & 0.8140 & 0.8300 & 0.6334 \\
 & Cognate & 1.0806 & 0.8083 & 0.8133 & 0.6190 & 1.0985 & 0.8225 & 0.8389 & 0.6431 \\
 & Features only & \colorbox{orange!20}{1.2491} & \colorbox{orange!20}{0.7086} & \colorbox{orange!20}{0.7080} & \colorbox{orange!20}{0.5146} & \colorbox{orange!20}{1.3339} & \colorbox{orange!20}{0.6825} & \colorbox{orange!20}{0.6893} & \colorbox{orange!20}{0.4986} \\
\midrule
\multirow{10}{*}{cn}
 & All features & \colorbox{green!10}{0.9594} & 0.8235 & 0.8200 & 0.6300 & \colorbox{green!30}{0.9681} & \colorbox{green!30}{0.8351} & \colorbox{green!30}{0.8428} & \colorbox{green!30}{0.6568} \\
 & Top-1/group & 0.9951 & 0.8258 & \colorbox{green!30}{0.8365} & \colorbox{green!30}{0.6493} & 0.9844 & 0.8139 & 0.8134 & 0.6215 \\
 & Top-3/group  & 0.9651 & \colorbox{green!10}{0.8278} & 0.8231 & 0.6330 & 1.0126 & 0.8217 & 0.8334 & 0.6456 \\
 & Frequency & \colorbox{green!30}{0.9492} & \colorbox{green!30}{0.8314} & \colorbox{green!10}{0.8298} & \colorbox{green!10}{0.6397} & 0.9899 & 0.8293 & 0.8355 & 0.6472 \\
 & Surface-form & 0.9857 & 0.8173 & 0.8165 & 0.6245 & 0.9840 & \colorbox{green!10}{0.8298} & \colorbox{green!10}{0.8419} & \colorbox{green!10}{0.6498} \\
 & MLM & 0.9649 & 0.8218 & 0.8205 & 0.6305 & \colorbox{green!10}{0.9796} & 0.8280 & 0.8403 & \colorbox{green!10}{0.6498} \\
 & Surprisal & 0.9689 & 0.8225 & 0.8218 & 0.6305 & 1.0089 & 0.8231 & 0.8351 & 0.6460 \\
 & Semantic & 1.0053 & 0.8050 & 0.8025 & 0.6102 & 1.0030 & 0.8261 & 0.8335 & 0.6433 \\
 & Cognate & 0.9693 & 0.8186 & 0.8164 & 0.6256 & 0.9807 & 0.8264 & 0.8357 & 0.6463 \\
 & Features only & \colorbox{orange!20}{1.1647} & \colorbox{orange!20}{0.7159} & \colorbox{orange!20}{0.7080} & \colorbox{orange!20}{0.5190} & \colorbox{orange!20}{1.2377} & \colorbox{orange!20}{0.6823} & \colorbox{orange!20}{0.6925} & \colorbox{orange!20}{0.5081} \\
\bottomrule
\end{tabular}%
}
\caption{Feature ablation results for Spanish, German, and Chinese. Top-1/group and Top-3/group are built by selecting the highest-ranked one or three features within each feature group according to Kendall correlation with the training GLMM score.\colorbox{green!30}{\strut best}, \colorbox{green!10}{\strut near-best}, \colorbox{orange!20}{\strut worst}.}
\label{tab:feature-ablation-kendall-short}
\end{table*}
\begin{table}[ht]
\centering
\small
\setlength{\tabcolsep}{4pt}
\renewcommand{\arraystretch}{0.95}
\begin{tabular}{llcc}
\toprule
\textbf{Language} & \textbf{Run} & \textbf{RMSE $\downarrow$} & \textbf{Pearson $\uparrow$} \\
\midrule
\multirow{3}{*}{es}
&\colorbox{green!30}{All features} & \colorbox{green!30}{1.132} & \colorbox{green!30}{0.813} \\
& Frequency & 1.134 & 0.808 \\
& Top-3/family & 1.140 & 0.813 \\
\midrule
\multirow{3}{*}{de}
& All features & 1.079 & 0.819 \\
& \colorbox{green!30}{Frequency} & \colorbox{green!30}{1.037} & \colorbox{green!30}{0.834} \\
& Surprisal & 1.078 & 0.818 \\
\midrule
\multirow{3}{*}{cn}
& All features & 0.930 & 0.856 \\
& \colorbox{green!30}{Frequency} & \colorbox{green!30}{0.891} & \colorbox{green!30}{0.860} \\
& Top-1/family & 0.919 & 0.858 \\
\bottomrule
\end{tabular}
\caption{Official closed-track test results for UOL@IDEM submissions. Green highlights the best run.}
\label{tab:uolidem-official-runs}
\end{table}
We analyse the engineered features in three steps. First, we compute Kendall's $\tau$ correlation between each feature and the gold \texttt{GLMM score} for each language. Second, we use these rankings to construct two compact feature sets: Top-1 per feature family, which keeps the single highest-ranked feature in each family, and Top-3 per feature family, which keeps the three highest-ranked features in each family. Third, we compare these compact settings with models trained using only one feature family at a time. This separates feature--target association from the predictive value of compact subsets and single feature families.

The Kendall rankings show that frequency and lexical familiarity are the most consistent individual signals. In Spanish, the top-ranked features are subtitle-based target frequency ($|\tau|=0.2730$), the retrieval prior combining clue/context support and candidate-space size ($|\tau|=0.2643$), and subtitle contextual diversity ($|\tau|=0.2572$). In German, the retrieval prior ranks first ($|\tau|=0.2985$), followed by subtitle frequency ($|\tau|=0.2837$) and subtitle contextual diversity ($|\tau|=0.2680$). In Chinese, the strongest associations are subtitle frequency ($|\tau|=0.4128$), subtitle contextual diversity ($|\tau|=0.4127$), and the KELLY lexical familiarity score ($|\tau|=0.4029$). 

Table~\ref{tab:feature-ablation-kendall-short} shows that that frequency-based features (subtitle frequency and contextual diversity) emerge as the most consistent and strongest individual predictors across languages, but its role differs by L1. Spanish is the clearest frequency-driven case: the frequency-only model gives the best Dev RMSE, 1.0855, slightly ahead of all features, 1.0952, with nearly identical OOF RMSE, 1.0987 versus 1.0988. German shows a more balanced pattern, where surprisal gives the best Dev RMSE, 1.0861, narrowly ahead of all features, 1.0873, while frequency, 1.0941, and cognate/form-similarity, 1.0985, remain close. Chinese has the most distributed signal: all features perform best, with Dev RMSE 0.9681, but MLM, cognate/form-similarity, lexical/surface form, and frequency all remain competitive. 

The official test results in Table~\ref{tab:uolidem-official-runs} partly confirm this pattern. The best UOL@IDEM runs achieve RMSE 1.132 for Spanish, 1.037 for German, and 0.891 for Chinese. Spanish is best with the all-features run, while German and Chinese are best with the frequency-oriented run. Thus, although richer feature combinations remain useful during development, frequency-based predictors generalise most robustly on the hidden test set for German and Chinese.

\subsection{Error analysis}
\label{sec:error-analysis}

The feature-importance and ablation results show that in addition to the baseline sentence encoder, the models learn a strong difficulty signal, especially from frequency and lexical familiarity. However, a strong ranking does not guarantee good calibration. We therefore examine where frequency becomes insufficient, whether predictions are compressed across the difficulty scale, and which feature patterns distinguish low-error from high-error cases.

We first examine calibration by dividing the development items into five quantile-based gold-difficulty bands. Band~1 contains the easiest items and Band~5 the hardest items. For each language and band, we compute RMSE, MAE, Kendall's $\tau$, and signed bias, defined as $\hat{y}-y$, where positive values indicate overprediction of difficulty \citep{kendall1938new}. Uncertainty is estimated using bootstrap resampling \citep{efron1994introduction}.

\begin{figure}[H]
    \centering
    \includegraphics[width=\columnwidth]{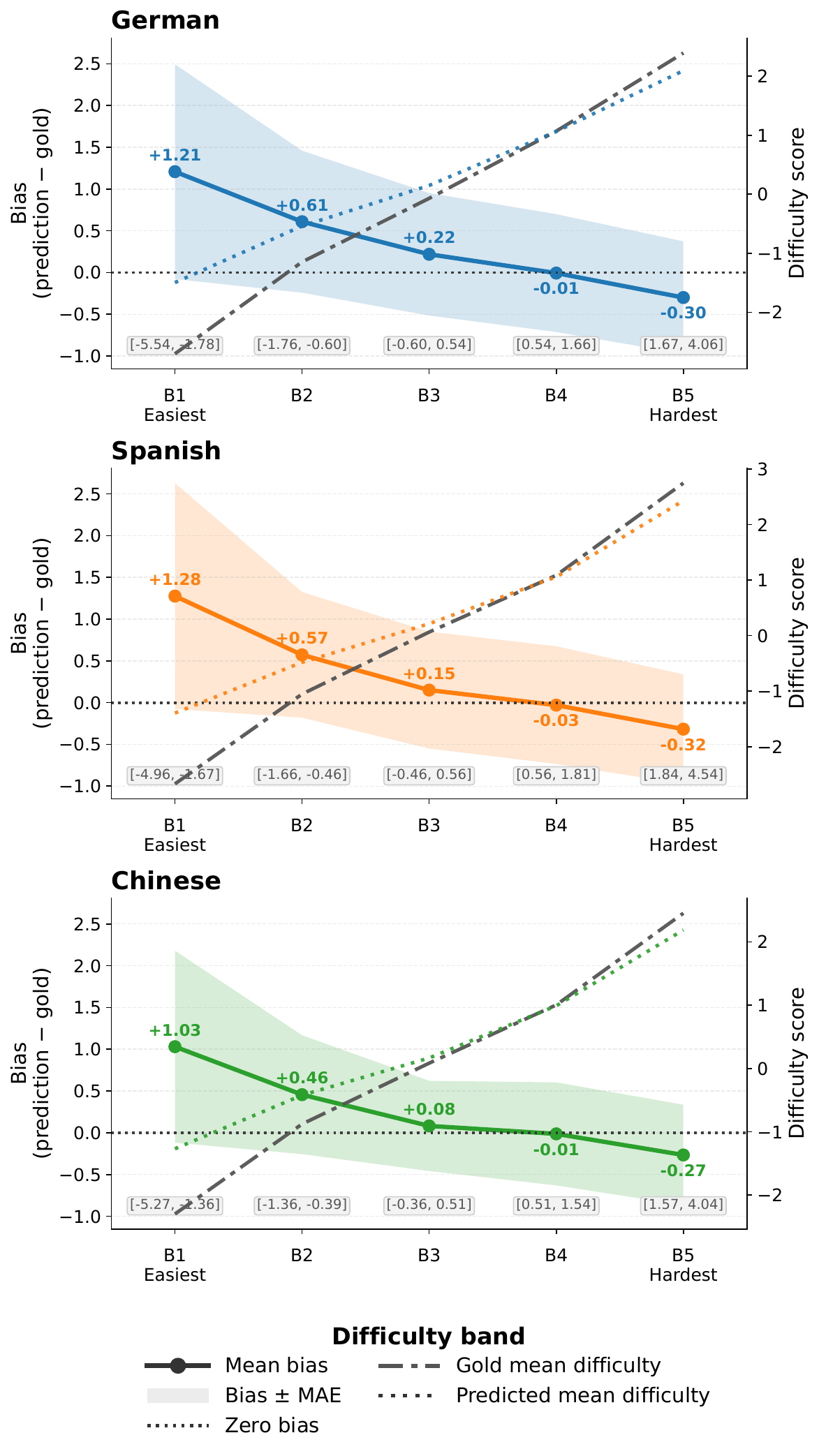}
  
   \caption{\small{
Calibration by gold difficulty band for German, Spanish, and Chinese. 
Bands B1--B5 are ordered from easiest to hardest according to the gold difficulty score. 
The solid line shows mean prediction bias, defined as predicted difficulty minus gold difficulty, and the shaded region shows bias $\pm$ MAE. 
Positive bias in the easiest bands indicates over-prediction of difficulty, while negative bias in the hardest band indicates under-prediction.
}}
    \label{fig:error-bias-band}
\end{figure}

Figure~\ref{fig:error-bias-band} shows that the language-specific neural fusion models broadly follow the gold difficulty scale, but with a clear compression toward the centre. This compression is visible in the solid mean-bias line, where bias is defined as predicted difficulty minus gold difficulty. In Band~1, which contains the easiest items, the bias is strongly positive: $+1.21$ for German, $+1.28$ for Spanish, and $+1.03$ for Chinese. This means that genuinely easy items are predicted as more difficult than they are. At the other end of the scale, Band~5 shows the opposite pattern: the bias is negative, with values of $-0.30$, $-0.32$, and $-0.27$, respectively, indicating that the hardest items are predicted as slightly easier than their gold scores. These two effects show that predictions are pulled away from the extremes and toward the middle of the difficulty scale. The same pattern is also reflected in the mean difficulty lines: the predicted mean difficulty line changes less sharply across bands than the gold mean difficulty line. Thus, the main weakness is not a failure to rank items by difficulty, but a calibration problem in which the model underestimates the full spread of the difficulty scale. The full error-analysis materials are available online.\footnote{\url{https://tinyurl.com/UoL-BEA-Error-Analysis}}

\begin{table}[ht]
\centering
\small
\setlength{\tabcolsep}{4pt}

\resizebox{\linewidth}{!}{%
\begin{tabular}{lrrrrr}
\toprule
\textbf{Lang.} & \textbf{RMSE} & \textbf{MAE} & \textbf{Bias} & \textbf{$\tau$} & \textbf{Band bias pattern} \\
\midrule
de  & 1.117 & 0.850 & +0.346 & 0.629 & $+1.21 \rightarrow -0.30$ \\
es  & 1.111 & 0.834 & +0.331 & 0.651 & $+1.28 \rightarrow -0.32$ \\
zh  & 0.975 & 0.724 & +0.258 & 0.648 & $+1.03 \rightarrow -0.27$ \\
\bottomrule
\end{tabular}%
}

\caption{
Overall development-set error profile. Bias is the mean signed error $\hat{y}-y$. The final column summarises the change from Band~1 to Band~5, showing scale compression in all three languages.
}
\label{tab:error-overview}
\end{table}

Table~\ref{tab:error-overview} summarises the same pattern at language level. Chinese has the lowest RMSE and MAE, but it still shows the same movement from overprediction in the easiest band to mild underprediction in the hardest band. This connects directly to the ablation results: frequency provides a robust global difficulty signal, but it can also pull predictions toward an average level of difficulty rather than fully adapting to the extremes of the learner-specific scale.
\begin{figure*}[t]
    \centering
    \includegraphics[
        width=0.92\textwidth]
     {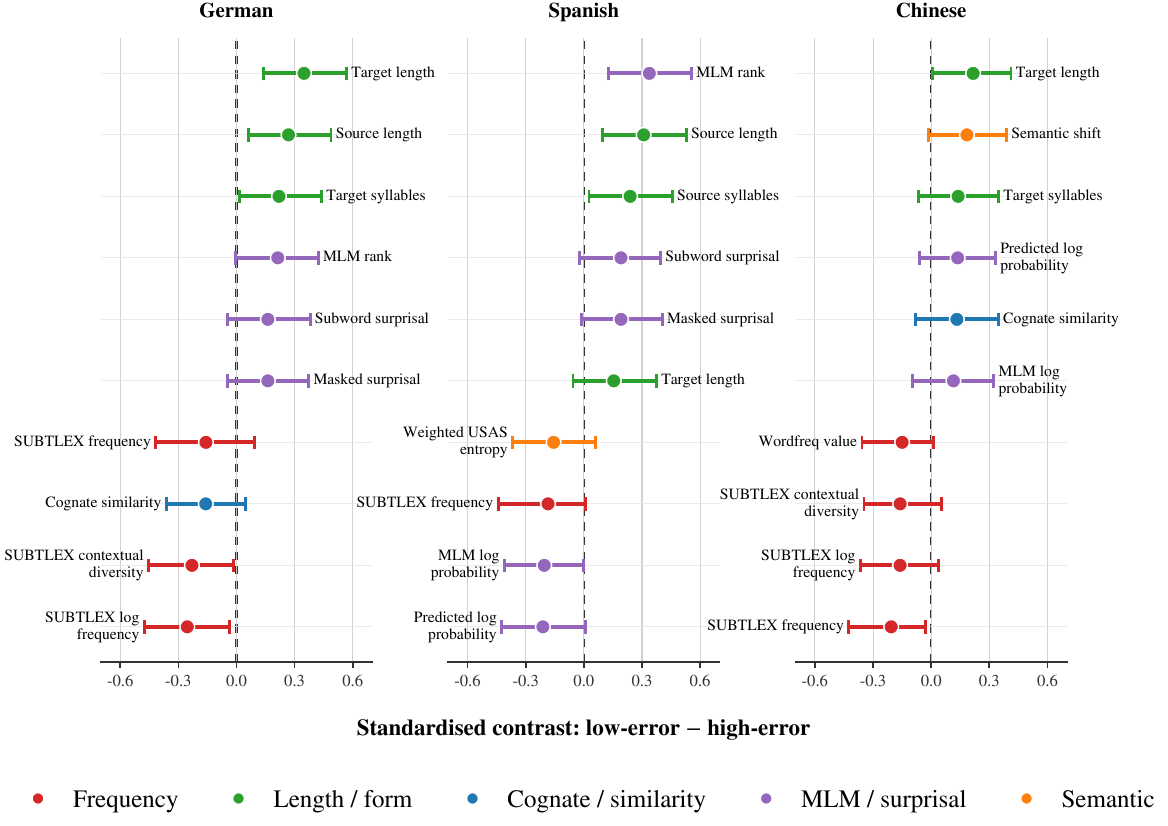}
    \caption{\small
    Matched low-error versus high-error contrasts within language and difficulty band.
    Positive values indicate features higher in low-error items; negative values indicate
    features higher in high-error items.
    }
    \label{fig:matched-low-high-contrasts}
\end{figure*}
Absolute-error associations with individual features are weak: most Kendall correlations are below $|\tau|=0.05$. The strongest cue is prediction entropy, $\mathtt{pred\_ent}$, with $\tau=0.048$, followed by MLM entropy, learner-frequency percentile, target length, semantic-domain entropy, prediction margin, and cognate similarity. Residual errors therefore appear when several cues diverge, such as frequency, contextual predictability, semantic ambiguity, form similarity, and L1--English transparency.

To examine these interactions more directly, we compare low-error and high-error items within the same language and difficulty band. This controls for broad gold difficulty: we are not simply comparing easy words with hard words, but asking what distinguishes better and worse predictions among items of comparable true difficulty.

Within each language--band group, we define low-error items as the lowest-error quartile and high-error items as the highest-error quartile. Since many features are skewed, ordinal, or tied, we use Mann--Whitney tests and Cliff's $\delta$ to compare feature distributions, and apply Benjamini--Hochberg correction to control the false discovery rate \citep{mann1947test,cliff1993dominance,benjamini1995controlling}. Figure~\ref{fig:matched-low-high-contrasts} reports the standardised contrast between the two groups, computed as low-error minus high-error. Positive values indicate features that are higher in better-predicted items; negative values indicate features that are higher in worse-predicted items.

The controlled contrasts clarify the source of the calibration errors. Low-error items are usually those where several cues point in the same direction: frequency, lexical form, semantic evidence, and contextual predictability jointly support the predicted difficulty. High-error items arise when these cues conflict, so the model receives mixed evidence about whether the item should be easy or difficult.

In German, low-error items show stronger form and contextual-predictability signals, including target length, source length, target syllable count, MLM rank, and surprisal features. High-error items are more associated with subtitle frequency, subtitle contextual diversity, and cognate similarity. This pattern indicates that German predictions are most stable when frequency is supported by contextual and form-based evidence, but less stable when frequency or surface similarity dominates without enough contextual support.

Spanish shows the clearest example of this conflict. The most error-prone cell is Spanish Band~1, where genuinely easy items are often overpredicted. Within this band, the strongest descriptive contrasts involve subword surprisal, MLM log probability, and pseudo-log-likelihood surprisal. The largest contrast is for $\mathtt{surp\_sub\_sum}$, with Cliff's $\delta=-0.298$ and a raw Mann--Whitney $p=0.031$; however, this effect does not remain significant after Benjamini--Hochberg correction ($p_{\mathrm{BH}}=0.522$), so we treat it as exploratory. The pattern is still informative: some easy items may be frequent, familiar, cognate-supported, or transparent for Spanish-speaking learners even when the immediate context does not make them highly predictable to an MLM-style model. The model then overpredicts difficulty because contextual predictability conflicts with learner-oriented familiarity.

Chinese has the strongest overall performance, but it follows the same mechanism. Low-error items combine form, semantic-shift, cognate-similarity, and probability-based support, including target length, target syllables, semantic shift, predicted log probability, cognate similarity, and MLM log probability. High-error items are more strongly associated with frequency features such as word-frequency value, subtitle contextual diversity, subtitle log-frequency, and subtitle frequency. Thus, even for Chinese, predictions are most reliable when frequency is reinforced by semantic and form-based evidence rather than operating as the main cue alone.

Overall, the error analysis refines the feature-ablation findings. Frequency is the strongest global cue and explains much of the high ranking performance, but it does not guarantee calibrated predictions. This explains why high Kendall correlations can coexist with systematic calibration errors: the models often rank items correctly, but their predicted scores are compressed toward the middle, leading to overprediction for easy items and underprediction for hard items.

\section{Conclusion}
\label{sec:conclusion}

This paper presented UOL@IDEM, a closed-track system for L1-aware English vocabulary difficulty prediction. The system combines multilingual contextual representations with engineered features capturing frequency, lexical surface form, surprisal, retrieval evidence, cognate-like similarity, and semantic alignment.

Across the development set, neural fusion with sentence-embedding encoders gives the strongest performance, improving over the closed-track baselines by 0.2618 RMSE for Spanish, 0.2407 for German, and 0.2069 for Chinese. On the official test set, the best submissions achieve RMSE scores of 1.132, 1.037, and 0.891 for Spanish, German, and Chinese, respectively.

The analysis shows that explicit linguistic features remain useful even with strong multilingual encoders. Frequency is the most stable signal, while form, surprisal, retrieval, cognate, and semantic features provide complementary evidence for L1-aware prediction. At the same time, the error analysis shows that strong ranking performance does not always translate into well-calibrated difficulty estimates, especially at the easiest and hardest ends of the scale.

Future work should focus on calibration and on more compact models that retain the most informative frequency, form, surprisal, and semantic cues while better modelling L1-specific transfer and contextual recoverability.

\section{Limitations}
\label{sec:limitations}

Our work is limited to the closed-track shared-task data, so we do not use external learner corpora, larger lexical resources, multilingual joint training, or large language models. These could be useful for rare targets, noisy source-word entries, and ambiguous source--target mappings.

The engineered features improve interpretability but add pipeline complexity, including preprocessing, frequency lookup, retrieval scoring, surprisal estimation, semantic tagging, and cognate/similarity scoring. Following the closed-track requirements, we train separate models for each L1, which does not explicitly model shared versus language-specific difficulty signals.

The main evaluation limitation is calibration. The models overpredict the easiest items and slightly underpredict the hardest ones, indicating regression toward the middle of the difficulty scale. This analysis is restricted to the development set because gold labels for the hidden test set are not available.
\section*{Ethics Statement}

This work uses data released for research under the shared-task conditions. The task concerns educational assessment and therefore has potential downstream implications for fairness and learner support. We do not claim that model predictions should replace psychometric validation or expert judgment; rather, they should be used as supportive signals in item development and educational NLP pipelines.

\section*{Acknowledgments}

We thank the BEA 2026 shared-task organisers for releasing the dataset and the evaluation framework. This document is part of a project that has received funding from the European Union's Horizon Europe research and innovation program under Grant Agreement No. 101132431 (iDEM Project).  The University of Leeds was funded by UK Research and Innovation (UKRI) under the UK government’s Horizon Europe funding guarantee (Grant Agreement No. 10103529).  The views and opinions expressed in this document are solely those of the author(s) and do not necessarily reflect the views of the European Union. Neither the European Union nor the granting authority can be held responsible for them.

\bibliography{custom}
\clearpage
\appendix
\onecolumn
\section{Feature Inventory}
\label{app:features}

This appendix summarises the engineered features used in our experiments.
Feature names follow the column names used in the analysis files and figures.
For readability, features are grouped by functional role rather than presented
as a single flat inventory.Table~\ref{tab:appendix_engineered_features} gives the full engineered-feature inventory.

\newcommand{\feat}[1]{\texttt{\small\seqsplit{#1}}}

\begingroup
\catcode`\_=12  

\begin{longtable}{
    @{}
    >{\raggedright\arraybackslash}p{5.4cm}
    >{\raggedright\arraybackslash}p{9.2cm}
    @{}
}
\caption{Engineered features grouped by functional role.}
\label{tab:appendix-engineered-features} \\  
\toprule
\textbf{Feature} & \textbf{Description} \\
\midrule
\endfirsthead

\multicolumn{2}{c}{\tablename~\thetable{} -- continued from previous page} \\
\toprule
\textbf{Feature} & \textbf{Description} \\
\midrule
\endhead

\midrule
\multicolumn{2}{r}{\textit{Continued on next page}} \\
\endfoot

\bottomrule
\endlastfoot

\multicolumn{2}{@{}l}{\textbf{Frequency}} \\
\midrule
\feat{freq_en}             & English frequency estimate for the target word. \\
\feat{freq_en_pct_rank}    & Percentile rank of the English target frequency. \\
\feat{sub_wf}              & SUBTLEX word-frequency measure. \\
\feat{sub_lg10wf}          & Log-scaled SUBTLEX word-frequency measure. \\
\feat{sub_cd}              & SUBTLEX contextual-diversity measure. \\
\feat{sub_lg10cd}          & Log-scaled SUBTLEX contextual-diversity measure. \\
\feat{kelly_rank}          & Rank of the target word in a learner-oriented lexical frequency resource. \\
\feat{kelly_pct}           & Percentile rank derived from the Kelly learner-frequency list. \\
\feat{kelly_points}        & Difficulty-oriented lexical score from the Kelly list. \\
\feat{wf_zipf}             & Zipf-scaled word-frequency score for the English target word. \\
\feat{wf_value}            & Raw or transformed corpus-based word-frequency value. \\

\addlinespace[6pt]
\multicolumn{2}{@{}l}{\textbf{Lexical, clue, and preprocessing}} \\
\midrule
\feat{tgt_len}                               & Number of characters in the English target word. \\
\feat{src_len}                               & Number of characters in the cleaned L1 source word. \\
\feat{tgt_syll}                              & Syllable count of the English target word. \\
\feat{target_pos}                            & Part-of-speech information for the English target word. \\
\feat{source_pos}                            & Part-of-speech information for the L1 source word, where available. \\
\feat{clue_len}                              & Length of the target clue. \\
\feat{clue_hidden_chars}                     & Number of hidden characters in the target clue. \\
\feat{clue_hidden_ratio}                     & Proportion of hidden characters in the target clue. \\
\feat{L1_source_word_raw}                    & Original unnormalised source-word field preserved during preprocessing. \\
\feat{L1_source_word_excluded_word}          & Word appearing inside an exclusion note in the raw source-word field. \\
\feat{L1_source_word_has_excluded_word}      & Indicator that the source-word field contains an exclusion note. \\
\feat{L1_source_word_excluded_note_type}     & Type of exclusion note detected during preprocessing. \\
\feat{L1_source_word_has_alternative}        & Indicator that multiple source-word candidates are present. \\
\feat{L1_source_word_alternative_count}      & Number of alternative source-word candidates detected. \\
\feat{L1_source_word_alternatives}           & Preserved list of source-word alternatives extracted during preprocessing. \\

\addlinespace[6pt]
\multicolumn{2}{@{}l}{\textbf{Retrieval}} \\
\midrule
\feat{gold_rank}                       & Rank position of the gold English target in the candidate list after retrieval and reranking. \\
\feat{gold_score}                      & Final reranked score assigned to the gold target. \\
\feat{top1_score}                      & Score of the highest-ranked retrieved candidate. \\
\feat{top2_score}                      & Score of the second-ranked retrieved candidate. \\
\feat{margin12}                        & Difference between the top-1 and top-2 candidate scores. \\
\feat{gold_cos}                        & Cosine similarity between the gold target and the retrieved representation. \\
\feat{gold_prob}                       & Probability-like score associated with the gold target under retrieval/reranking. \\
\feat{nn_pred_cos}                     & Cosine similarity associated with the nearest-neighbour prediction. \\
\feat{nn_pred_prob}                    & Probability-like confidence score for the nearest-neighbour prediction. \\
\feat{nn_entropy}                      & Entropy of the retrieved candidate distribution. \\
\feat{cand_count}                      & Number of valid candidate targets considered after filtering. \\
\feat{retrieval_target_prior}          & Prior retrieval score for the target independent of full contextual reranking. \\
\feat{retrieval_target_in_context}     & Context-sensitive retrieval score for the target. \\
\feat{baseline_pred_matches_target}    & Indicator that the retrieval baseline predicts the correct target. \\
\feat{baseline_clue_overlap}           & Degree of overlap between the predicted candidate and the target clue. \\
\feat{baseline_pred_len}               & Length-based compatibility between the retrieved candidate and the target constraint. \\

\addlinespace[6pt]
\multicolumn{2}{@{}l}{\textbf{MLM and surprisal}} \\
\midrule
\feat{mlm_logp}               & Mean masked-language-model log probability assigned to the gold target wordpieces. \\
\feat{mlm_rank}               & Rank of the gold target under the masked-language-model prediction. \\
\feat{mlm_ent}                & Entropy of the masked-token distribution. \\
\feat{mlm_top1_match}         & Indicator that the top MLM prediction matches the gold target. \\
\feat{pred_logp}              & Log probability of the top predicted token or predicted target sequence. \\
\feat{pred_ent}               & Entropy associated with the prediction distribution. \\
\feat{pred_margin}            & Margin between the best and second-best MLM predictions. \\
\feat{mlm_target_pred_cos}    & Cosine similarity between gold-target and predicted-token embeddings. \\
\feat{surp_masked}            & Surprisal computed from simultaneous masking of all target wordpieces. \\
\feat{surp_pll}               & Surprisal computed using pseudo-log-likelihood with one masked wordpiece at a time. \\
\feat{surp_chain}             & Surprisal computed sequentially via a chain-rule decomposition over target subwords. \\
\feat{surp_sub_sum}           & Sum of surprisal contributions across target subwords. \\
\feat{surp_sub_mean}          & Mean surprisal across target subwords. \\

\addlinespace[6pt]
\multicolumn{2}{@{}l}{\textbf{Semantic and domain shift}} \\
\midrule
\feat{usas_domain_match}       & Indicator of semantic-domain compatibility between source and target. \\
\feat{usas_ent_unw}            & Unweighted entropy over possible semantic-domain assignments. \\
\feat{usas_ent_wtd}            & Weighted entropy over semantic-domain assignments. \\
\feat{sem_shift}               & Soft semantic-shift score measuring divergence between the L1 source word and the English target. \\
\feat{semantic_domain_overlap} & Degree of overlap between source- and target-side semantic domains. \\

\addlinespace[6pt]
\multicolumn{2}{@{}l}{\textbf{Cognate and cross-lingual form}} \\
\midrule
\feat{lev_sim}          & Weighted Levenshtein-based similarity between the L1 source word and the English target. \\
\feat{cog_sim}          & Cross-lingual form-similarity score used as a cognate or transparency cue. \\
\feat{char_ngram_sim}   & Character n-gram similarity between the L1 source word and the English target, where available. \\
\feat{cognet_match}     & Indicator or score derived from external cognate-link evidence, where available. \\
\caption{Engineered features grouped by functional role.}
\label{tab:appendix_engineered_features} \\
\end{longtable}
\endgroup

\FloatBarrier

\section{Neural-fusion model settings}
\label{app:neural-fusion-settings_sec}

\begin{table}[ht]
\centering
\scriptsize
\begin{tabular}{p{3.0cm}p{4.2cm}p{7.4cm}}
\toprule
\textbf{Category} & \textbf{Setting} & \textbf{Value} \\
\midrule

\multirow{4}{*}{Input and model}
& Text fields
& \texttt{L1\_context}, \texttt{L1\_source\_word}, \texttt{en\_target\_word}, optional \texttt{en\_target\_clue} \\

& Fusion model
& Enhanced neural fusion \\

& Encoder backbones
& XLM-R, mBERT, LaBSE, multilingual-E5, BGE-M3, and related multilingual encoders \\

& Text pooling
& Attention pooling or CLS pooling, depending on the model configuration \\

\midrule

\multirow{2}{*}{Architecture}
& Tabular branch
& Feed-forward network with configurable hidden layers, dropout, and optional residual connections \\

& Prediction head
& MLP regression head over the concatenated text and tabular representations \\

\midrule

\multirow{9}{*}{Optimisation}
& Optimiser
& AdamW \\

& Encoder learning rate
& $2 \times 10^{-5}$ \\

& Task-head learning rate
& $10^{-3}$ \\

& Weight decay
& 0.01 \\

& Warmup ratio
& 0.1 \\

& Schedule
& Linear warmup and decay \\

& Gradient clipping
& Maximum norm 1.0 \\

& Mixed precision
& Used when GPU support is available \\

& Checkpoint selection
& Validation RMSE \\

\midrule

\multirow{4}{*}{Training protocol}
& Epochs
& 5 \\

& Training batch size
& 16 \\

& Evaluation batch size
& 16 or 32, depending on the experiment configuration \\

& Early stopping
& Patience of 2 epochs \\

\midrule

\multirow{3}{*}{Evaluation}
& Cross-validation
& 5-fold \\

& Main metric
& RMSE \\

& Other metrics
& Pearson, Spearman, Kendall's $\tau$ \\

\bottomrule
\end{tabular}
\caption{Model, training, and optimisation settings used in the neural-fusion experiments.}
\label{tab:neural-fusion-settings}
\end{table}

\FloatBarrier

The Huber loss used in some optimisation settings is defined in Eq.~\ref{eq:huber-loss}.

\begin{equation}
L_\delta(a) =
\begin{cases}
\frac{1}{2}a^2 & \text{if } |a| \le \delta, \\
\delta\left(|a| - \frac{1}{2}\delta\right) & \text{otherwise.}
\end{cases}
\label{eq:huber-loss}
\end{equation}

\FloatBarrier

\section{USAS semantic-shift examples}
\label{app:usas-shift}

\newcommand{\usascell}[2]{%
\begin{minipage}[t]{\linewidth}
\raggedright
{\ttfamily\footnotesize #1}\par
\emph{#2}
\end{minipage}
}

\begin{table}[H]
\centering
\scriptsize
\renewcommand{\arraystretch}{1.25}

\begin{tabularx}{\textwidth}{
  >{\raggedright\arraybackslash}p{2.1cm}
  >{\raggedright\arraybackslash}X
  >{\raggedright\arraybackslash}X
  >{\centering\arraybackslash}p{0.9cm}
  >{\centering\arraybackslash}p{0.9cm}
}
\toprule
\textbf{Example}
& \textbf{Source USAS tags and categories}
& \textbf{Target USAS tags and categories}
& \textbf{Soft sim.}
& \textbf{Soft shift} \\
\midrule

\multicolumn{5}{@{}l}{\textbf{Operation: exact tag overlap}} \\
\addlinespace[2pt]

\textit{firme}--\textit{solid}
& \usascell{O4.1|O4.5}
  {General appearance and physical properties; Texture}
& \usascell{A5.1+|A5.3+|O1.1|O4.5|T2++}
  {Evaluation: Good/bad; Evaluation: Accuracy; Substances and materials: Solid; Texture; Time: Beginning and ending}
& 0.67 & 0.33 \\

\addlinespace[2pt]

\textit{riesgo}--\textit{risk}
& \usascell{A1.4|A15-|I1|I2|I2.1.3}
  {Chance/luck; Safety/Danger; Money generally; Business; Business: Generally}
& \usascell{A1.4|A15-}
  {Chance/luck; Safety/Danger}
& 0.78 & 0.22 \\

\midrule
\multicolumn{5}{@{}l}{\textbf{Operation: same fine-grained prefix}} \\
\addlinespace[2pt]

\textit{oyente}--\textit{listener}
& \usascell{S2.1|S2.2|S5|X2.3}
  {People: Female; People: Male; Groups and affiliation; Learn}
& \usascell{S2mf|X3.2}
  {People; Sensory: Sound}
& 0.69 & 0.31 \\

\addlinespace[2pt]

\textit{noticias}--\textit{news}
& \usascell{Q1|Q4}
  {Linguistic actions, states and processes; The Media}
& \usascell{Q4.2|Q4.3|T3-|X2.2+}
  {The Media: Newspapers; The Media: TV, radio and cinema; Time: Old/new/young; Knowledge}
& 0.60 & 0.40 \\

\midrule
\multicolumn{5}{@{}l}{\textbf{Operation: same major semantic field}} \\
\addlinespace[2pt]

\textit{ladrillo}--\textit{brick}
& \usascell{O1|O4.3|O4.4}
  {Substances and materials generally; Colour and colour patterns; Shape}
& \usascell{H1|O2|S8+}
  {Architecture and kinds of houses/buildings; Objects generally; Helping/hindering}
& 0.43 & 0.57 \\

\addlinespace[2pt]

\textit{debatir}--\textit{discuss}
& \usascell{A6.1|G1.2|Q1.1}
  {Comparing: Similar/different; Politics; Linguistic actions and communication}
& \usascell{Q2.1|Q2.2}
  {Speech: Communicative; Speech acts}
& 0.42 & 0.58 \\

\midrule
\multicolumn{5}{@{}l}{\textbf{Operation: related major semantic fields}} \\
\addlinespace[2pt]

\textit{sue\~no}--\textit{dream}
& \usascell{B3|T1.3}
  {Medicines and medical treatment; Time: Period}
& \usascell{A5.2-|X4.1|X7+}
  {Evaluation: True/false; Mental object: Conceptual object; Wanting/planning/choosing}
& 0.18 & 0.82 \\

\addlinespace[2pt]

\textit{obra}--\textit{comedy}
& \usascell{A1.1.1|I3.1}
  {General actions, making etc.; Work and employment: Generally}
& \usascell{E4.1+|K4|Q4.3}
  {Happy/sad: Happy; Drama, theatre and showbusiness; The Media: TV, radio and cinema}
& 0.20 & 0.80 \\

\midrule
\multicolumn{5}{@{}l}{\textbf{Operation: unrelated major semantic fields}} \\
\addlinespace[2pt]

\textit{fuegos}--\textit{fireworks}
& \usascell{L1|O1|O4.6}
  {Life and living things; Substances and materials generally; Temperature}
& \usascell{G3|K1}
  {Warfare, defence and weapons; Entertainment generally}
& 0.05 & 0.95 \\

\addlinespace[2pt]

\textit{circulaci\'on}--\textit{traffic}
& \usascell{O4.4}
  {Shape}
& \usascell{G2.1-|I2.2|M1|M3|M4|S1.1.2+}
  {Crime, law and order; Business: Selling; Moving/coming/going; Vehicles and transport on land; Shipping/swimming; Reciprocity}
& 0.05 & 0.95 \\

\midrule
\multicolumn{5}{@{}l}{\textbf{Operation: coverage gap}} \\
\addlinespace[2pt]

\textit{extensi\'on}--\textit{spread}
& \usascell{Z99}
  {Unmatched}
& \usascell{A2.1|A6.3+|F1|N3.3|N3.6|N5+|Q4.2}
  {Affect: Modify/change; Comparing: Variety; Food; Measurement: Distance; Measurement: Area; Quantities; The Media: Newspapers}
& -- & 0.85$^{\dagger}$ \\

\bottomrule
\end{tabularx}

\caption{Spanish examples of the soft USAS semantic-shift measure using richer source and target tag sets. Each cell reports the full available USAS tag set, with the corresponding category names underneath. The operation row describes the strongest source--target tag-pair relation used to characterise each example group, while the soft similarity and soft shift are computed over the full source and target tag sets. Therefore, an item can contain an exact overlapping tag but still receive a non-zero shift score when the remaining tags differ. $^{\dagger}$Coverage-gap scores are penalty values rather than observed semantic-distance scores.}
\label{tab:soft-usas-shift-examples}
\end{table}

\FloatBarrier

\end{document}